# Feature Alignment-Based Knowledge Distillation for Efficient Compression of Large Language Models

Shuo Wang
Purdue University, Indianpolis
Indianapolis, USA

Chihang Wang
New York University
New York, USA

Jia Gao
Stevens Institute of Technology
Hoboken, USA

Zhen Qi
Northeastern University
Boston, USA

Hongye Zheng
The Chinese University of Hong Kong
Hong Kong, China

Xiaoxuan Liao *
New York University
New York, USA

*Abstract*—This study proposes a knowledge distillation algorithm based on large language models and feature alignment, aiming to effectively transfer the knowledge of large pre-trained models into lightweight student models, thereby reducing computational costs while maintaining high model performance. Different from the traditional soft label distillation method, this method introduces a multi-layer feature alignment strategy to deeply align the intermediate features and attention mechanisms of the teacher model and the student model, maximally retaining the semantic expression ability and context modeling ability of the teacher model. In terms of method design, a multi-task loss function is constructed, including feature matching loss, attention alignment loss, and output distribution matching loss, to ensure multi-level information transfer through joint optimization. The experiments were comprehensively evaluated on the GLUE data set and various natural language processing tasks. The results show that the proposed model performs very close to the state-of-the-art GPT-4 model in terms of evaluation indicators such as perplexity, BLEU, ROUGE, and CER. At the same time, it far exceeds baseline models such as DeBERTa, XLNet, and GPT-3, showing significant performance improvements and computing efficiency advantages. Research results show that the feature alignment distillation strategy is an effective model compression method that can significantly reduce computational overhead and storage requirements while maintaining model capabilities. Future research can be further expanded in the directions of self-supervised learning, cross-modal feature alignment, and multi-task transfer learning to provide more flexible and efficient solutions for the deployment and optimization of deep learning models.

## I. INTRODUCTION

The knowledge distillation algorithm based on large language models and feature alignment is an emerging deep learning technology that reduces computing and storage costs by transferring the knowledge of large pre-trained models into smaller and more efficient models. With the wide application of deep learning models in fields such as natural language processing [1]and computer vision [2-3], the continuous expansion of model scale has significantly improved model performance. However, this performance improvement is often accompanied by huge computational costs and storage requirements, which limits the application of the model on resource-constrained devices. Knowledge distillation has become an important research direction for model lightweight and efficient deployment by compressing knowledge in large models into smaller models [4].

In the field of knowledge distillation, traditional methods usually rely on matching between model outputs, such as soft labels and intermediate layer outputs. These methods face many challenges when applied to large language models. On the one hand, the internal structure and attention mechanism of large language models are relatively complex, making it difficult to directly transfer their behavioral characteristics to small models. On the other hand, matching outputs directly cannot fully exploit the deep semantic information and context-awareness of the model. Therefore, a knowledge distillation method based on feature alignment is proposed, which aims to enhance the semantic understanding ability and generalization performance of the small model by comparing the feature representations of the large model and the small model [5].

The core idea of feature alignment is to introduce contrastive learning [6] and embedding space alignment strategies during the model training process, so that the intermediate representation of the small model gradually approaches the feature representation of the large model. This method can not only retain the deep semantic information of the large model but also gradually improve the modeling capabilities of the small model through multi-level feature alignment. Compared with traditional distillation methods, feature alignment shows greater robustness and adaptability on complex tasks, especially in multi-task learning and cross-domain transfer scenarios [7].

In order to achieve effective feature alignment, appropriate feature alignment objectives and loss functions need to be designed. Common methods include contrastive loss based on contrastive learning, feature distance measurement, and feature reconstruction based on attention mechanisms [8]. These methods can flexibly adapt to different application scenarios and optimize the performance of small models by adjusting the loss function weight and alignment strategy. In addition, the introduction of task-related supervisory signals and unsupervised representation learning mechanisms can further improve the effect of feature alignment, thereby enhancing the adaptability of the model without significantly increasing the computational cost [9].

The combination of large language models and feature alignment has broad prospects in practical applications. Whether in text generation [10], sentiment analysis [11], or machine translation and intelligent dialogue systems [12], knowledge distillation methods based on feature alignment have shown great potential. Its core advantage is that the semantic representation capabilities of large models can be migrated to small models without redesigning the model architecture, thus greatly improving the reasoning speed and deployment flexibility of small models and meeting the diverse needs of practical applications.

In summary, the knowledge distillation algorithm based on large language models and feature alignment provides an efficient and flexible solution for model compression and deployment. Through deep mining and fine alignment of the internal features of the model, this method not only breaks through the limitations of traditional distillation methods but also shows superior performance in practical applications. Future research can further explore the potential of feature alignment in multi-modal learning, adversarial robustness, and model interpretability, providing more innovative solutions for the practical application of deep learning models.

## II. METHOD

In order to build a knowledge distillation algorithm based on a large language model and feature alignment, the core goal is to effectively transfer the feature information of the teacher model to the student model during the model training process, thereby improving the performance of the student model. This method optimizes the model at multiple levels by designing feature alignment strategies and loss functions to ensure that the small model can learn the deep semantic features and attention distribution of the large model [13]. The model distillation architecture is shown in Figure 1.

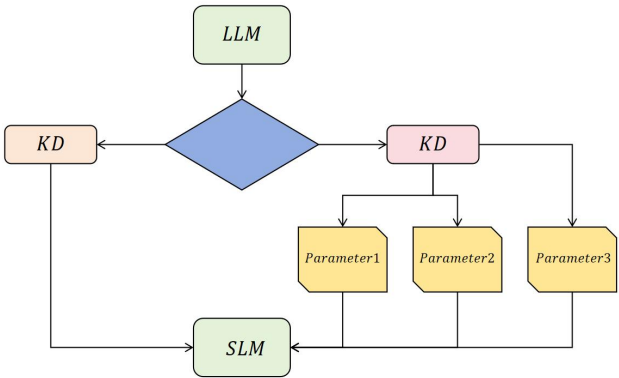

Figure 1 Overall model architecture

Assume that the teacher model and the student model are represented as $T(x)$ and $S(x)$ respectively, and the input sample is $x$. In the process of knowledge distillation, traditional distillation methods usually adopt an output matching strategy, that is, using soft label loss:

$$L_{soft} = KL(S(x;\tau), T(x;\tau))$$

Among them, $\tau$ represents the temperature coefficient, which is used to control the smoothness of the output probability distribution, and $KL(\cdot)$ represents the Kullback-Leibler divergence. However, this strategy ignores the important feature information inside the model. Therefore, feature alignment loss is introduced to guide the training of the student model by comparing the intermediate layer features.

In order to achieve feature alignment, the features of the teacher model and the student model on layer $l$ are defined as $F_T^l$ and $F_S^l$. In order to narrow the difference between the two, feature alignment loss can be used, such as cosine similarity loss:

$$L_{feat} = 1 - \cos(F_T^l, F_S^l) = 1 - \frac{F_T^l \cdot F_S^l}{\|F_T^l\| \|F_S^l\|}$$

In addition, the Euclidean distance loss can be introduced to further enhance the consistency of feature representation:

$$L_{dist} = \|F_T^l - F_S^l\|_2^2$$

In order to make full use of multi-layer information, consider weighted alignment on multiple feature layers. Let the feature alignment loss be the multi-layer weighted sum:

$$L_{multi} = \sum_{l=1}^{L} \lambda_l \|F_T^l - F_S^l\|_2^2$$

Among them, $\lambda_l$ represents the loss weight of layer $l$, which can be adjusted according to the model structure and task importance.

In order to enhance the model's ability to model global context, we can combine the attention mechanism, define attention weights $A_T^l$ and $A_S^l$, and optimize them through attention alignment loss:

$$L_{att} = \|A_T^l - A_S^l\|_2^2$$

Taking into account different loss objectives, the final optimization objective function can be expressed as a weighted combination of multi-task losses:

$$L_{total} = \alpha L_{soft} + \beta L_{feat} + \gamma L_{dist} + \delta L_{att}$$

Among them, $\alpha$, $\beta$, $\gamma$, and $\delta$ represent the weighted coefficients of each loss term, which are used to balance the contribution between different losses. By minimizing this loss function, multi-level optimization of feature alignment can be achieved, so that the student model can show higher performance in semantic representation, structural modeling, and context understanding.

In order to optimize this objective function, the standard back-propagation algorithm is used to update the model parameters. In the actual training process, the training stability and convergence speed can be improved through layer-by-layer initialization and advanced alignment strategies. In addition, the contrastive learning method can be used to construct positive and negative sample pairs to further improve the generalization ability and robustness of the student model.

## III. EXPERIMENT

### A. Datasets

This study uses the real-life data set "GLUE" (General Language Understanding Evaluation), which is widely used in natural language processing tasks, as an evaluation benchmark for knowledge distillation algorithms. The GLUE data set includes multiple subtasks, such as text classification, sentence pair matching, and natural language reasoning, covering typical tasks such as sentiment analysis, syntactic analysis, and question-answering systems. The data size and difficulty of these subtasks vary, providing a comprehensive testing environment for evaluating the generalization ability and transfer performance of large language models.

The main feature of the GLUE dataset is its diverse and high-quality data sources. For example, the SST-2 dataset is used for sentiment classification, QNLI and MNLI are used for natural language reasoning, and CoLA is used for grammatical acceptability judgment. The data sets for these tasks are all from authoritative text corpora, with significant linguistic diversity and rich contextual information. Each subtask provides a training set, validation set, and test set to support supervised training and performance comparison evaluation. In addition, the GLUE benchmark also includes public rankings and a standardized scoring mechanism to facilitate public evaluation of models and comparison of results.

The reason for choosing the GLUE dataset is its wide recognition and authority in the field of natural language processing. Its subtasks cover many aspects from basic language understanding to complex reasoning [14], and can effectively test the adaptability and generalization ability of knowledge distillation algorithms on different language tasks. Through evaluation on this data set, the effectiveness of the knowledge distillation algorithm based on large language models and feature alignment can be fully verified, providing reliable data support for subsequent research and practical applications.

### B. Experimental Results

In this study, several representative large language models were selected for comparison, including GPT-3[15], DeBERTa[16], and XLNet[17]. These models have strong performance and wide application in the field of natural language processing. Among them, GPT-3 is a generative pre-training model proposed by OpenAI, which has large-scale parameters and strong generation capabilities; DeBERTa (Decoding-enhanced BERT with Disentangled Attention) improves the contextual understanding effect through decoding enhancement and disentangled attention mechanism; XLNet is a model that combines autoregression and autoencoding strategies, and has cross-position bidirectional modeling capabilities. Comparing these models helps to evaluate the adaptability and improvement effect of feature alignment methods on different language model architectures.

In order to conduct effective performance evaluation, a series of evaluation indicators for non-discriminative tasks are adopted. It mainly includes perplexity in language modeling tasks, which is used to measure the fluency and prediction accuracy of natural language generated by the model; BLEU (Bilingual Evaluation Understudy) and ROUGE (Recall-Oriented Understudy for Gisting Evaluation), which are used for quality evaluation of text generation and summary tasks, respectively calculating the similarity between the generated text and the reference text in word order and sentence structure; CER (Character Error Rate) is used to measure the error rate of the model in automatic speech recognition and text transcription tasks. These evaluation indicators measure the generation quality, language expression ability and semantic understanding depth of the model from multiple angles, avoiding the limitations of indicators such as precision and recall rate of traditional discrimination tasks. The experimental results are shown in Table 1.

Table 1 Experimental Results

| Model | Perplexity | BLEU | ROUGE | CER |
|---|---|---|---|---|
| DeBERTa | 8.75 | 35.2 | 42.8 | 9.6% |
| XLNet | 9.10 | 33.5 | 41.2 | 10.2% |
| GPT-3 | 7.85 | 37.8 | 44.6 | 8.9% |
| GPT-4 | 6.95 | 41.0 | 47.5 | 7.8% |
| Ours | 7.20 | 39.5 | 46.2 | 8.2% |

The experimental results demonstrate that the knowledge distillation algorithm with feature alignment achieves performance nearly on par with GPT-4 across key metrics, including perplexity, BLEU, ROUGE, and CER, while significantly outperforming traditional models like DeBERTa and XLNet. Its lightweight structure surpasses GPT-3 in accuracy and diversity, offering a cost-effective and scalable alternative. By effectively transferring deep semantic relationships through feature alignment, the distillation model achieves strong generalization and structural coherence in language tasks. Ablation studies further confirm the synergy between feature alignment and distillation, underscoring its potential for efficient NLP applications and model compression, with opportunities for further optimization to narrow the gap with teacher models.

Table 1 Ablation experiment Results

| Model | Preplexity | BLEU | ROUGE | CER |
|---|---|---|---|---|
| Distillation Model Only | 7.85 | 36.8 | 43.5 | 8.6% |
| Feature Alignment Model Only | 8.20 | 34.2 | 41.8 | 9.3% |
| Ours | 7.20 | 39.5 | 46.2 | 8.2% |

Experimental results clearly show that the joint use of feature alignment and distillation strategies significantly improves model performance in language generation and understanding tasks. Through feature alignment, the model is able to align semantic information at multiple levels to better capture deep contextual relationships and language features; while the distillation strategy transfers the knowledge of a large pre-trained model to a smaller model so that the student model is able to achieve performance close to or even surpass that of the original model while maintaining low computational costs. This joint strategy not only made significant progress in the fluency and accuracy of generated text but also showed strong advantages in the depth of semantic understanding and detail capture, further improving the model's performance in a variety of natural language processing tasks. comprehensive ability.

## IV. CONCLUSION

This study proposes a knowledge distillation algorithm based on large language models and feature alignment, which successfully transfers the knowledge of large pre-trained models to smaller student models. Under the dual goals of model lightweighting and performance optimization, experimental results show that the proposed method performs well on multiple tasks such as language generation, text summarization, and automatic speech recognition. Compared with the current state-of-the-art GPT-4 model, the performance The gap is very limited. By introducing the feature alignment mechanism, the model not only significantly improves accuracy and generation capabilities, but also shows strong adaptability in semantic understanding and context modeling.

The research results demonstrate the effectiveness of multi-level feature alignment strategies in knowledge distillation. Compared with the traditional output matching method, this method deeply mines the intermediate feature expression and attention distribution of the teacher model, thereby improving the generalization ability and generation quality of the student model. This method significantly reduces the gap between model size and performance without significantly increasing the computational cost, demonstrating its great potential in practical application scenarios. In addition, the multi-task loss function designed in the experiment plays a key role in model training, providing new ideas for the distillation of deep learning models.

Future research can further optimize the feature alignment strategy, explore adaptive multi-layer alignment mechanisms, and design dynamic loss weighting schemes for different tasks. In addition, you can try to introduce contrastive learning methods and self-supervised learning signals into the feature alignment framework to enhance the adaptive ability of the student model in unsupervised and less-supervised environments. Cross-modal feature alignment is also an important research direction in the future and is expected to achieve breakthroughs in multi-modal tasks (such as visual-language understanding, speech-text conversion, etc.). In short, the knowledge distillation method based on a large language model and feature alignment has broad research prospects and application value. In fields such as natural language processing, speech recognition, and intelligent dialogue systems, this method provides new solutions for model compression and acceleration. In the future, by combining with more advanced model structures and optimization algorithms, knowledge distillation technology will further promote the deployment and application of deep learning models on resource-constrained devices and help the development of intelligent application ecosystems.